\documentclass{article}

\usepackage{microtype}
\usepackage{graphicx}
\usepackage{subfigure}
\usepackage{booktabs} 

\usepackage{hyperref}

\usepackage[accepted]{icml2025}

\usepackage{amsmath}
\usepackage{amssymb}
\usepackage{mathtools}
\usepackage{amsthm}
\usepackage{mathabx}

\usepackage[capitalize,noabbrev]{cleveref}

\theoremstyle{plain}
\newtheorem{theorem}{Theorem}[section]

\theoremstyle{definition}
\newtheorem{definition}[theorem]{Definition}

\theoremstyle{remark}

\usepackage[textsize=tiny]{todonotes}
\usepackage{multirow}

\icmltitlerunning{Reducing Smoothness with Provably More Expressive Memory Enhanced Hierarchical Graph Neural Networks}

\begin{document}

\twocolumn[

\icmltitle{Reducing Smoothness with Expressive Memory Enhanced \\Hierarchical Graph Neural Networks}



\icmlsetsymbol{equal}{*}

\begin{icmlauthorlist}

\icmlauthor{Thomas Bailie}{auckland}
\icmlauthor{Yun Sing Koh}{auckland}
\icmlauthor{S. Karthik Mukkavilli}{mercuria}
\icmlauthor{Vara Vetrova}{canterbury}

\end{icmlauthorlist}

\icmlaffiliation{auckland}{School of Computer Science, The University of Auckland, New Zealand}
\icmlaffiliation{mercuria}{Mercuria, Geneva}
\icmlaffiliation{canterbury}{Department of Computer Science, University of Canterbury, New Zealand}

\icmlcorrespondingauthor{Thomas Bailie}{tbai869@aucklanduni.ac.nz}

\icmlkeywords{Machine Learning, ICML}

\vskip 0.3in]



\printAffiliationsAndNotice{} 

\begin{abstract}

Graphical forecasting models learn the structure of time series data via projecting onto a graph, with recent techniques capturing spatial-temporal associations between variables via edge weights. Hierarchical variants offer a distinct advantage by analysing the time series across multiple resolutions, making them particularly effective in tasks like global weather forecasting, where low-resolution variable interactions are significant. A critical challenge in hierarchical models is information loss during forward or backward passes through the hierarchy. We propose the Hierarchical Graph Flow (HiGFlow) network, which introduces a memory buffer variable of dynamic size to store previously seen information across variable resolutions. We theoretically show two key results: HiGFlow reduces smoothness when mapping onto new feature spaces in the hierarchy and non-strictly enhances the utility of message-passing by improving Weisfeiler-Lehman (WL) expressivity. Empirical results demonstrate that HiGFlow outperforms state-of-the-art baselines, including transformer models, by at least an average of 6.1\% in MAE and 6.2\% in RMSE. Code is available at \url{https://github.com/TB862/HiGFlow.git}.

\end{abstract}

\section{Introduction}




Graphical machine learning, which leverages the inherent topology of graph structures, has numerous applications in the physical sciences \cite{karniadakis2021physics, wu2022graph, cuomo2022scientific, xiong2021graph}. Global weather forecasting, for instance, involves predicting a signal distributed across Earth's surface using time-varying simulated or observational data \cite{lam2023graphcast, bauer2015quiet}. In this domain, Graph Neural Networks effectively model teleconnections at fixed spatial-temporal resolutions \cite{yi2024fouriergnn}. However, interactions between climate variables occur at a large scale across multiple spatial-temporal resolutions, leading to suboptimal modelling of underlying global dynamics. 




Real-world graphs are, furthermore, often locally dense with significant bottlenecks, making it challenging for message-passing techniques to propagate information \cite{di2023over, alon2020bottleneck} effectively. Hierarchical methods mitigate these issues by summarising global trends through the aggregation of related variables into a single descriptor \cite{wang2018multilevel, zhang2017stock}. In a graphical context, a clustering algorithm maps node groupings to super nodes, forming a coarse graph that captures non-local relationships in the time series \cite{oskarsson2024probabilistic, yang2021hierarchical}. However, current models struggle to preserve low-resolution information from top hierarchical levels due to challenges in mapping between feature-spaces of differing dimensionalities, reducing their descriptive power. This issue arises from two key factors: (1) the backward pass through the hierarchy fails to retain relational information between nodes and super nodes, leading to discordant low-resolution information when passing back down the hierarchy, and (2) reliance on unlearnable mappings to project clusters onto super nodes, causing to overfitting to non-coarse signals that {\it provably} increases feature space smoothness.

We propose the Hierarchical Graph Flow (HiGFlow) network, a novel hierarchical graph framework which addresses these issues. HiGFlow incorporates a memory buffer variable that remains consistent across all hierarchical resolutions. As a result, HiGFlow can propagate information from previously observed resolutions, functioning similarly to the hidden state in an LSTM \cite{hochreiter1997long} for long temporal rollouts. Intuitively, this allows for greater information retention, increasing the effectiveness of stacking hierarchical layers. HiGFlow furthermore incorporates node embeddings into learnable mappings between nodes and super nodes, provably enhancing the use of contextual information when mapping across spaces of differing dimensionalities. 

We study theoretically the effects of HiGFlow's increased accessibility to multi-resolution information. Specifically, we prove three theoretical claims; \textbf{(i)} linear transition functions smooth the resulting feature space, thereby diminishing the effectiveness of subsequent operations \cite{rusch2023survey}; \textbf{(ii)} models using highly non-linear transition functions, {\it e.g.} HiGFlow, limit smoothness, and \textbf{(iii)} when using a memory-buffer variable, adding hierarchical depth non-strictly increases expressivity under the Weisfeiler-Lehman (WL) test \cite{xu2018powerful}, making HiGFlow more expressive than a $1$-WL hash function.


Our experimental results demonstrate that HiGFlow outperforms state-of-the-art forecasting models, including both graph-based models \cite{yi2024fouriergnn} and transformer-based approaches \cite{nie2022time,liu2023itransformer} on all datasets. In addition, we show via sensitivity tests that increased non-linearity in mappings between node and super-node preserves model performance in a practical setting. Our contributions are as follows: 

\begin{itemize} 
    \item We introduce a novel information retention mechanism that captures multi-resolution spatial-temporal relationships in a memory buffer, enhancing the effectiveness of stacking hierarchical levels.
    
    \item We prove that neural networks decrease smoothness of feature-spaces in the hierarchy, while linear mappings strictly increase it, and further empirically demonstrate this finding.
    
    \item We prove that HiGFlow, via the memory-buffer, non-strictly increases WL expressivity when stacking hierarchical layers, surpassing a $1$-WL hash function. Empirical results show HiGFlow outperforming state-of-the-art models.   
\end{itemize}

\section{Preliminaries}

We outline within this section essential concepts and definitions required for the rest of this work. Our framework is theoretically grounded in message-passing on a time-series graphical embedding.  

\subsection{Forecasting}

We define a {\it time series} at time $t$ as signal $\mathbf{x}(t \vert T) \in \mathbb{R}^{N \times T}$; a total of $N$ spatial variables across a sliding window of $T$ time steps, starting at $t$. The signal then exists over the interval $[t, t+T_{in}]$. The forecasting task takes an input signal $\mathbf{x}(t\vert T_{in})$ with buffer size $T_{in}$, and predicts the immediate future signal over $T_{out}$ time steps $\mathbf{x}(t + T_{in}\vert T_{out})$. 

In our work, we denote the $T$ dimensional row-vector for variable $i$ as $\mathbf{x}_i(t\vert T)$, and to ease notation, we additionally use $\mathbf{x}(t)$ to mean $\mathbf{x}(t \vert 0) \in \mathbb{R}^{N \times 1}$.

\subsection{Graph Embeddings}

We enrich the descriptiveness of the feature space by embedding our input signal onto a hidden dimension of size $D$ with weight vector $\mathbf{w}_{e} \in \mathbb{R}^{1\times D}$ and activation function $\sigma$:
\begin{equation}
    \mathbf{h}(t|T_{in}) = \sigma(\mathbf{w}_{e} \cdot \mathbf{x}(t\vert T_{in})),
\end{equation}
where we now have $\mathbf{h}(t|T_{in}) \in \mathbb{R}^{(N\times T_{in})\times D}$. 

As shown in \cite{cao2020spectral} and later \cite{yi2024fouriergnn}, we learn the implicit spatial-temporal relationships within $\mathbf{h}(t \vert T_{in})$. That is, we consider each $(i, t) \in [1,N] \times [1,T_{in}]$ to be a node within the vertex set of graph $G$, and find the weight of an edge between any two nodes via  multi-headed self-attention \cite{vaswani2017attention}; $G$ describes spatial-temporal relationships.
We enforce a hard constraint on variable relationships by truncating edge weights below a fixed threshold $\tau \in (0,1]$.


\subsection{Message-Passing Graph Neural Networks}

Message Passing Neural Networks (MPNNs) aggregate feature-vectors based on causal relationships described by the graph topology \cite{kipf2016semi, xu2018powerful, hamilton2017inductive, velickovic2018graph}. Subsequently, a new representation of the feature vectors is found.

The general message-passing equation is defined using the aggregation function $\text{AGG}$ and update function $\text{UP}$, which together map input $\mathbf{x}(t)$ to output $\mathbf{y}(t)$:
\begin{equation}
    \mathbf{y}_i(t) = \text{UP}\left(\mathbf{h}_i(t), \text{AGG}\left(\{\!\{ \mathbf{h}_j(t) \mid j \in \mathcal{N}(i) \}\!\}\right)\right),
    \label{eq:1agg}
\end{equation}
where $\{\!\{\}\!\}$ denotes a multi-set, and $\mathcal{N}(i)$ represents the set of nodes in the 1-hop neighbourhood of node $i$.



\begin{figure}
    \centering
    \includegraphics[width=0.9\columnwidth]{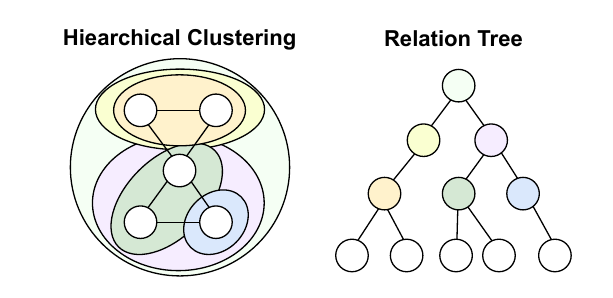}
    \caption{Hierarchical clustering on a graph and the corresponding relationships between regions within the graph.}
    \label{fig:clustering}
\end{figure}

\section{Hierarchical Graph Flow Network}

We introduce the Hierarchical Graph Flow (HiGFlow) network that, by iteratively coarsening an initial graph, views spatial-temporal interactions at multiple resolution scales. The framework incorporates learnable embedding and lifting maps that, according to Theorem \ref{th:th2}, do not increase the smoothness of the resulting feature spaces. In contrast, we prove previous methods \cite{cini2023graph, guo2021hierarchical} increase smoothness; see Theorem \ref{th:th1}. With the introduction of a persistent memory buffer variable, we construct a prediction with information at a mixture of resolutions.

\subsection{Framework Overview}

\begin{figure*}
    \centering
    \includegraphics[width=2.05\columnwidth, height=55mm]{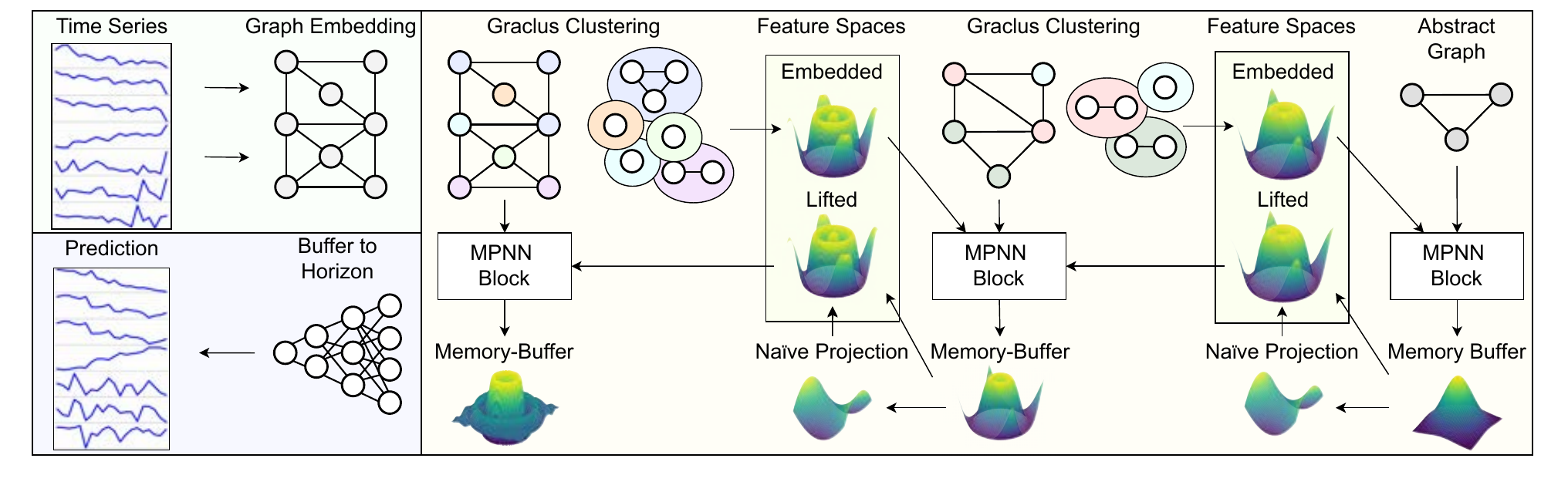}
    \caption{The framework begins by embedding a time series onto a graph, where edges represent relative intra-series associations. Nodes are greedily clustered based on edge weights. Information at any resolution is handled by embedding cluster feature vectors into a lower-resolution space or lifting lower-resolution data upward. A memory-buffer variable, incorporating the time-series topology, iteratively builds the prediction. Finally, a neural network maps the memory buffer to the prediction horizon.}
    \label{fig:main_techniqual}
\end{figure*}

Figure \ref{fig:main_techniqual} illustrates our overall approach. We model the topology of $\mathbf{x}(t \vert T_{in})$, coarsening its graph embedding via clique assignment to find intra-series dependencies across $K$ progressively lower resolutions. See Figure \ref{fig:clustering} for an illustration. 

\begin{definition}[Abstract and Predecessor Graphs]
    A graph $G_i = (V(G_i), E(G_i), W(G_i))$, consisting of a vertex set $V(G_i)$, an edge set $E(G_i)$, and edge weights $W(G_i)$, is called the \textit{predecessor} of an \textit{abstract} graph $G_{i+1}$ if $G_{i+1}$ is obtained by clustering on $V(G_i)$. In particular, if there exists a subset $S \subseteq V(G_i)$ that maps to a unique clique $m \in V_{i+1}$.
\end{definition}

We define a transition function as a mapping between the feature spaces of a predecessor and its abstraction, in either direction:

\begin{definition}[Transition Function] 
    A \textit{transition function} $ F: \mathcal{X}(G_n) \to \mathcal{X}(G_m) $ maps the feature space of $G_n$ to that of $G_m$. When $ n = i+1$ and $ m = i $, $ F $ is referred to as a \textit{lifting map}. Conversely, when $ n = i $ and $ m = i+1 $, $ F $ is called an \textit{embedding map}. For $ j \in V(G_n)$ and subgraph $ C_{j, m} $ of $ G_m $, the following holds:
    \begin{equation}
        \mathbf{x}_{j,m}(t) = F(C_{j, n}).
    \end{equation}
\end{definition}
In other hierarchical forecasting settings, such as those in \cite{cini2023graph, guo2021hierarchical}, the feature space over abstract graphs is derived using a straightforward statistical mapping. In the following theorem, we demonstrate that when the transition function is linear, the Dirichlet energy decreases, resulting in a loss of information due to increased smoothness in the feature space.
\begin{definition}[Dirichlet Energy]
    The {\it Dirichlet energy} measures the similarity between feature-vectors within $G_i$:
    \begin{equation}
        \mathcal{D}(G_i) = \sum_{(u, v) \in E_i}\Big\Vert \frac{\mathbf{x}_{u,i}}{d_{u,i}} - \frac{\mathbf{x}_{v,i}}{d_{v,i}}\Big\Vert_2,
    \end{equation}
    where $d_{u,i} = deg(u)$ is the degree of $u$ in $G_i$.
\end{definition}
\begin{theorem}
   The statistical transition function $M(C_{j,n}) = \sum_{u\in C_{j,n}} \mathbf{x}_{u,n}$ contracts the total Dirichlet energy when mapping from $\mathcal{X}(G_n)$ to $\mathcal{X}(G_m)$. In particular, for any $j \in V(G_{n})$, then it is the case that the strong condition holds:
    \begin{equation}
        \label{eq:eq2}
        \mathcal{D}(C_{j, m}) \leq  
        \sum_{u\in V(C_{j,n})} \mathcal{D}(C_{u, n}).
    \end{equation}
    \label{th:th1}
\end{theorem}
We prove Theorem \ref{th:th1}. in Appendix \textbf{A}. An increase in feature space smoothness is analogous to information loss, reducing the utility of subsequent operations \cite{rusch2023survey}. Our framework introduces \textbf{learnable} embedding and lifting maps to transition between predecessor and abstract graphs, consequently enhancing the expressivity of message-passing on the graph. We define $\mathbf{h}{:,i}(t)$ and $\mathbf{u}{:,i}(t)$ as the features at depth $i$, derived through the embedding and lifting mappings, respectively. During the backward pass, we store information from all resolutions in a depth-persistent memory buffer, $\mathbf{y}_{:,i}(t)$. This mechanism enables information reuse and functions similarly to the gated behaviour of a GRU \cite{chung2014empirical}.

\subsection{Embedding Nodes to Super Nodes}

\label{sec:embedding}

Embedding $G_i$ onto its abstraction involves three steps: {\bf (1)} partitioning the nodes of $G_i$ into a collection of cliques $\mathcal{C}$ which disjointly cover $V(G_i)$, and gathering their corresponding feature-vectors from $\mathbf{h}_{:,i}(t)$, {\bf (2)} aggregating and then mapping the gathered feature-vectors onto a super-node for each cluster, and {\bf (3)} learning a new topological structure using self-attention based on the feature space of the abstract graph.

{\bf Clustering.} We construct the collection $\mathcal{C}$ via spatial-temporal variable similarity inferred through the self-attention mechanism, emphasising plausible feature relationships rather than the feature vectors directly. We implement Graclus clustering \cite{dhillon2007weighted}, a greedy algorithm that selects nodes based on the highest edge weight, ensuring each cluster is simply connected.

{\bf Dimension Reduction.} We map information contained within $C_j$ via aggregating feature vectors onto a single global representation. To ensure diversity in the feature-space, we post-process the aggregation via a neural network $\mathbf{M}_{\phi_i}$ with parameters $\phi_i$. Specifically, $\mathbf{M}_{\phi_i}$ enables the feature space $\mathcal{X}(G_{i+1})$ to become learnable. Our embedding is written as: 
\begin{equation}
    \begin{split}
        \mathbf{\bar{h}}_{j,i+1}(t) &= \sum_{q \in C_j} \mathbf{h}_{q,i}(t),
    \end{split}
    \label{eq:hagg}
\end{equation}
\begin{equation}
    \begin{split}
        \mathbf{h}_{:,i+1}(t) &= \mathbf{M}_\phi\left(\mathbf{\bar{h}}_{:,i+1}(t) + \mathbf{W}_e^T\cdot\mathbf{v}\right).\\
    \end{split}
    \label{eq:hmap}
\end{equation}
Here, multiplication of node indices $\mathbf{v} = (1, \dots, {N_{i+1}})^T$ by the linear map $\mathbf{W}_e\in\mathbb{R}^{N_{i+1} \times N_{i+1}}$ enables a topology-free linear embedding of the nodes of $G_{i+1}$, which assists in establishing a vague notion of locality within $\mathbf{\bar{h}}_{j,i+1}(t)$. Our implementation of $\mathbf{M_{\phi_i}}$ transforms features only along the spatial dimension, significantly reducing the number of learnable parameters.  

Theorem \ref{th:th2} shows that a highly non-linear transition function can come arbitrarily close to not smoothing the feature space, {\it e.g.}, when $\mathbf{M}_{\phi_i}$ is a neural network, considering the universality theorem \cite{hornik1989multilayer}. We defer the proof and additional details, to Appendix {\bf A}.

\begin{theorem}
   Consider the truncated power-series expansion $p_B(\mathbf{\bar{x}}) = \sum^B_{b=1} \omega_b \cdot \overline{\mathbf{x}}^{(b)}$ of the function $f_\theta(\overline{\mathbf{x}})$, which contains $B$ non-zero power-series coefficients $\omega_b$. For statistical transition function $M(C) = \sum_{\mathbf{x} \in C} \mathbf{x}$, the non-linear transition function $F_\theta = f_\theta \circ M$ satisfies:
    \begin{equation}
        \left\vert \mathcal{D}(C_{j, i+1}) - \sum_{u \in C_{j, i+1}}\mathcal{D}(C_{u, i}) \right\vert \mapsto 0 \text{ as } B \mapsto \infty. \label{eq:assert}
    \end{equation}
    \label{th:th2}
\end{theorem}

{\bf Graph Topology.} We account for changing feature relationships introduced by the mappings in Equations \ref{eq:hagg} and \ref{eq:hmap}, such as strong correlations between nodes not captured by the edge weights or clique topology of the predecessor. Specifically, we recompute the graph topology using a new self-attention module.

\subsection{Lifting Super Nodes to Nodes}

\label{sec:lifting}

Lifting reconstructs $\mathcal{X}(G_{i})$ using the coarse upstream signal from low hierarchical levels. Consequently, the lifting operation is essential for mixing up and downstream signals $\mathbf{u}_{:,i}(t)$ and $\mathbf{h}_{:,i}(t)$ at hierarchical level $i$. Additionally, we introduce the {\it memory-buffer} $\mathbf{y}_{:,i}(t)$, a buffer that stores information from all previous abstract graph feature spaces. Unlike $\mathbf{u}_{:,i}(t)$, $\mathbf{y}_{:,i}(t)$ is a constrained with respect to the topology of all $G_j$, for $j <i$. Note that Theorem \ref{th:th2} also applies to lifting maps, as its proof in Appendix \textbf{A} does not require the unique assignment of a node to a cluster. 

{\bf Na\"ive Encoding.} To reconstruct the predecessor feature space from lower-resolution sources, a learnable encoding mitigates biases introduced during projection to a higher-dimensional space. The lifting map $\mathbf{L}_\gamma$, parameterised by $\gamma$, projects $\mathbf{y}_{:,i-1}(t) \in \mathbb{R}^{N_{i-1}}$ onto $\mathbf{e}_{:,i}(t) \in \mathbb{R}^{N_i}$:  
\begin{equation}
    \mathbf{e}_{:,i}(t) = \mathbf{L}_\gamma(\mathbf{y}_{:,i-1}(t)).
\end{equation}

{\bf Inverse Cluster Map.} We define the lifting transition function with parameters $\xi$ as $\mathbf{U}_\xi$. This function reconstructs the feature space $\mathcal{X}(G_i)$ using signals from lower levels in the hierarchy, in contrast to the embedding map, which processes information from higher levels. Collectively, $\mathbf{U}_\xi$ enables the integration of mixed-resolution contextual information within the hierarchical framework. The lifting operation is expressed as:
\begin{equation}
    \begin{split}
        \mathbf{u}_{:,i}(t) &= \mathbf{U}_{\xi}(\mathbf{y}_{:,i-1}(t) + \mathbf{W}^T_l\cdot\mathbf{v'}) + \mathbf{e}_{:,i}(t).\\
    \end{split}
\end{equation}
Here, $\mathbf{v'} = (1, \dots, N_i)^T$ and $\mathbf{W}_l \in\mathbb{R}^{N_i\times N_{i}}$ a linear embedding. 

\subsection{Memory-Buffer for Information Retention}

We compute $\mathbf{y}_{:,i}(t)$ using an MPNN applied to $G_i$, where our choice of feature vectors integrates information across all topological resolutions.

\textbf{Message Passing.} We exploit the expressivity of MPNNs to propagate hidden states to higher resolutions by combining $\mathbf{h}_{:,i}(t)$ and $\mathbf{u}_{:,i}(t)$. The updated output state $\mathbf{y}_{:,i}(t)$ is defined as:
\begin{equation}
    \mathbf{y}_{:,i}(t) = \text{MP}(\mathbf{h}_{:,i}(t) \vert\vert \mathbf{u}_{:,i}(t)),\,
    \label{eq:MP}
\end{equation}
where $\text{MP}$ represents the ubiquitous 1-hop message-passing framework from Equation \ref{eq:1agg}, and $\vert\vert$ denotes concatenation. The framework's final output is $\mathbf{y}_{:,1}(t)$. Theorem \ref{th:th3} demonstrates the effectiveness of including a memory buffer:
\begin{theorem}
    For a hierarchical graph framework with a memory buffer across $N$ levels, the memory-enabled hash function $g_N(x_1, \dots, x_N) = h_N(x_1, g_{N-1}(x_2, \dots, x_N))$, where $h_N$ is an injective function, induces a graph colouring $\mathcal{X}_{N:1}(G)$ satisfying:
    \[
        \Big\vert \mathcal{X}^t_{N:1}(G) \Big\vert \geq \Big\vert \mathcal{X}^t_{N-1:1}(G) \Big\vert \geq \dots \geq \Big\vert \mathcal{X}^t_{1:1}(G) \Big\vert
    \]
    \label{th:th3}
\end{theorem}
In this context, $\mathcal{X}^t_{1:1}(G)$ is the colouring of $G$ under some hash-function $h$ bounded by the $1$-WL isomorphism test. Theorem \ref{th:th3} demonstrates that storing information from all resolutions in a memory-buffer non-strictly increases the expressiveness of any hash function, an effect that stacks with hierarchical depth. We provide a proof of Theorem \ref{th:th3} in Appendix \textbf{B}.  

\begin{figure}
    \centering
    \includegraphics[width=\columnwidth]{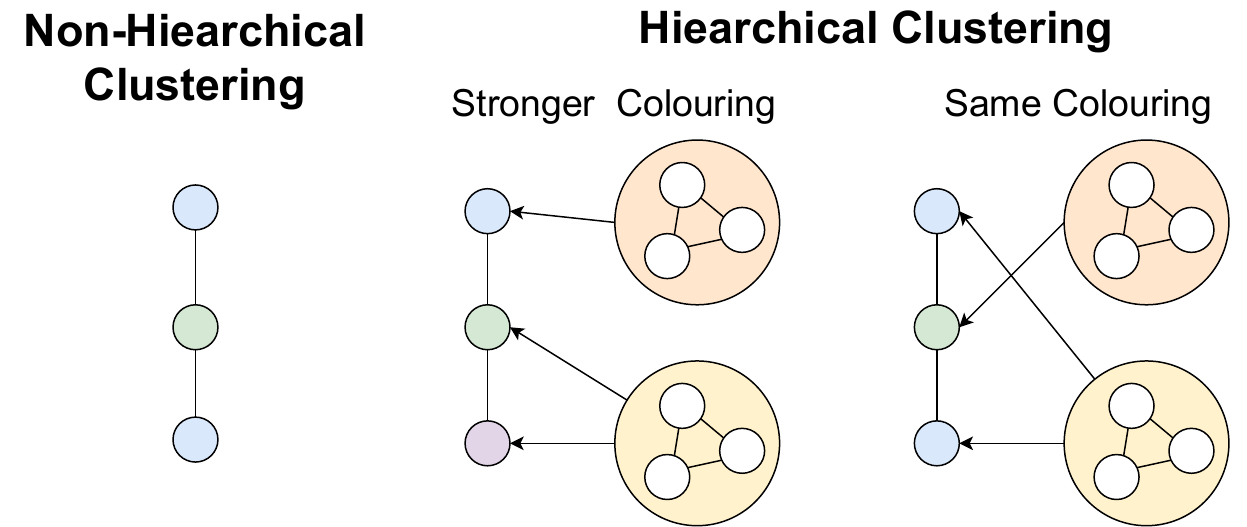}
    \caption{The benefit of a hierarchical framework when using a memory buffer. Hash function mappings of nodes with a cluster colour as an auxiliary argument are conditionally unique.}
    \label{fig:enter-label}
\end{figure}

\section{Experiments}

We evaluate the HiGFlow network against state-of-the-art forecasting models on eight real-world datasets and conduct sensitivity and ablation tests. Appendix {\bf C} contains further sensitivity tests. 

\subsection{Experimental Set Up}

{\bf Datasets.} We select datasets from a diverse range of real-world forecasting tasks, including Electricity, ECG, Solar, and Traffic. Additionally, we use the ETTm1, ETTm2, ETTh1, and ETTh2 datasets from \citet{wu2022graph}. We split all datasets are split into 60\%, 20\%, and 20\% for training, validation, and testing, respectively, with z-score normalisation.

{\bf Baselines.} We compare our HiGFLow network with state-of-the-art Fourier transform orientated forecasting models, including FourierGNN \cite{yi2024fouriergnn}, FreTS, NLinear and DLinear \cite{yi2024frequency}. Additionally, we evaluate against recent graphical forecasting models, such as StemGNN \cite{cao2020spectral}, and classic methods; LSTM \cite{hochreiter1997long}, and GRU \cite{chung2014empirical}. Given the well-established forecasting capabilities of transformer models, our analysis incorporates recent transformer-based architectures, such as Informer \cite{zhou2021informer} and Autoformer \cite{wu2022graph}. We further include state-of-the-art transformer baselines from  \citet{liu2023itransformer}; iTransformer, based on the Transformer model \cite{vaswani2017attention}, and iFlashformer, an improvement on Flashformer \cite{hua2022transformer}.

{\bf Experiment Settings.} We run our experiments using a single A100 NVIDIA GPU. We train all models using the Mean Squared Error Loss function and RMSProp optimiser, with all baseline model training parameters fixed to those used in \citet{cao2020spectral}. For HiGFlow, we tuned the learning rate on the ETTh1 dataset, changing it from $10^{-4}$ to $5\cdot10^{-4}$. We report the Mean Absolute Error (MAE) and Root Mean Squared Error (RMSE). As in \cite{cao2020spectral, yi2024fouriergnn}, we set the default input and prediction buffer sizes to $3$ and $12$ respectively. When implementing HiGFlow, we choose StemGNN \cite{cao2020spectral} blocks as the message-passing mapping of Equation \ref{eq:MP}.

\subsection{Evaluation}

\begin{table*}[ht]
\centering
\caption{Comparison of MAE and RMSE across different datasets and models.}
\resizebox{\textwidth}{!}{%
\begin{tabular}{l|cc|cc|cc|cc|cc|cc|cc|cc}
\hline
Models          & \multicolumn{2}{c|}{ECG} & \multicolumn{2}{c|}{Solar} & \multicolumn{2}{c|}{Traffic} & \multicolumn{2}{c|}{Electricity} & \multicolumn{2}{c|}{ETTh1} & \multicolumn{2}{c|}{ETTh2} & \multicolumn{2}{c|}{ETTm1} & \multicolumn{2}{c}{ETTm2} \\
                & MAE  & RMSE              & MAE  & RMSE              & MAE  & RMSE             & MAE  & RMSE               & MAE  & RMSE          & MAE  & RMSE          & MAE  & RMSE          & MAE  & RMSE          \\\hline
LSTM            &  0.745 & 1.051           & 0.075 & 0.165            & 0.330 & 0.637          & 0.341    & 0.492        & 0.717 & 1.001       & 0.333 & 0.439      & 0.253 & 0.385      & 0.151 & 0.202  \\
GRU             &  0.733 & 1.039           & 0.074 & 0.162            & 0.333 & 0.632          & 0.342    & 0.485        & 0.656 & 0.927      & 0.380 & 0.452       & 0.236 & 0.370       & 0.136 & 0.186       \\
StemGNN         & 0.638 & 0.895            & 0.065 & 0.226            & 0.273 & 0.540           & 0.326 & 0.475            & 0.396 & 0.582      & 0.311 & 0.391   & 0.210 & 0.347 &  0.115 & 0.165  \\
FourierGNN      & 0.644 & 0.909            & 0.051 & 0.128            & 0.278 & 0.551             & 0.266 & 0.398              & 0.375 & 0.567      & 0.165 & 0.232       & 0.202 & 0.341       & 0.104 & 0.157       \\
FreTS           & 0.642 & 0.899            & 0.053 & 0.136            & 0.320 & 0.613            & 0.334  & 0.501            & 0.379 & 0.575      & 0.169 & 0.235       & 0.202 & 0.342      & 0.106 & 0.158       \\
DLinear         & 0.712 & 0.976            & 0.115 & 0.219            & 0.465 & 0.774        & 0.381    & 0.549       & 0.415 & 0.605      & 0.173 & 0.238       & 0.203 & 0.345       &  0.104 & 0.158       \\
NLinear         & 0.752 & 1.040            & 0.068 & 0.158            & 0.319 & 0.613          & 0.410    & 0.602        & 0.436 & 0.644      & 0.165 & 0.232       & 0.206 & 0.354       & 0.104 & 0.158       \\
Autoformer        & 0.715 & 0.987 & 0.073 & 0.161 & 0.345 & 0.589 & 0.277 & 0.395 & 0.383 & 0.567 & 0.176 & 0.245 & 0.202 & 0.338 & 0.103 & 0.154 \\
Informer          & 0.623 & 0.891 & 0.065 & 0.154 & 0.312 & 0.608 & 0.265 & 0.374 & 0.398 & 0.581 & 0.190 & 0.256 & 0.247 & 0.380 & 0.134 & 0.183 \\
iFlashformer    &  0.717 &  1.010 & 0.063 & 0.149 & 0.283 & 0.557 & 0.288 & 0.415 & 0.349 & 0.532 & 0.156 & 0.219 & 0.215 & 0.359 & 0.115 & 0.162 \\
iTransformer      & 0.641 & 0.908 & 0.049 & 0.129 & 0.298 & 0.600 & 0.303 & 0.467 &  0.364 & 0.567 & 0.158 & 0.224 & 0.202 & 0.344 & 0.105 & 0.159 \\ \midrule
HiGFlow(1)        & 0.648 & 0.910            & 0.055 & 0.133            & 0.305  & 0.593            & 0.347 & 0.524         & 0.391 & 0.596      & 0.173 & 0.239       & 0.204 & 0.347 & 0.108 & 0.160      \\
HiGFlow(*)        & {\bf 0.621} & {\bf 0.889} & {\bf 0.046} & {\bf 0.120} & {\bf 0.265} & {\bf 0.534} & {\bf 0.217} & {\bf 0.359} & {\bf 0.349} & {\bf 0.532} & {\bf 0.156} & {\bf 0.219} & {\bf 0.190} & {\bf 0.338} & {\bf 0.102} & {\bf 0.153}\\
\bottomrule
\end{tabular}}
\label{tab:results}
\end{table*}

\begin{figure*}[!htbp]
    \centering
    \includegraphics[width=1.825\columnwidth]{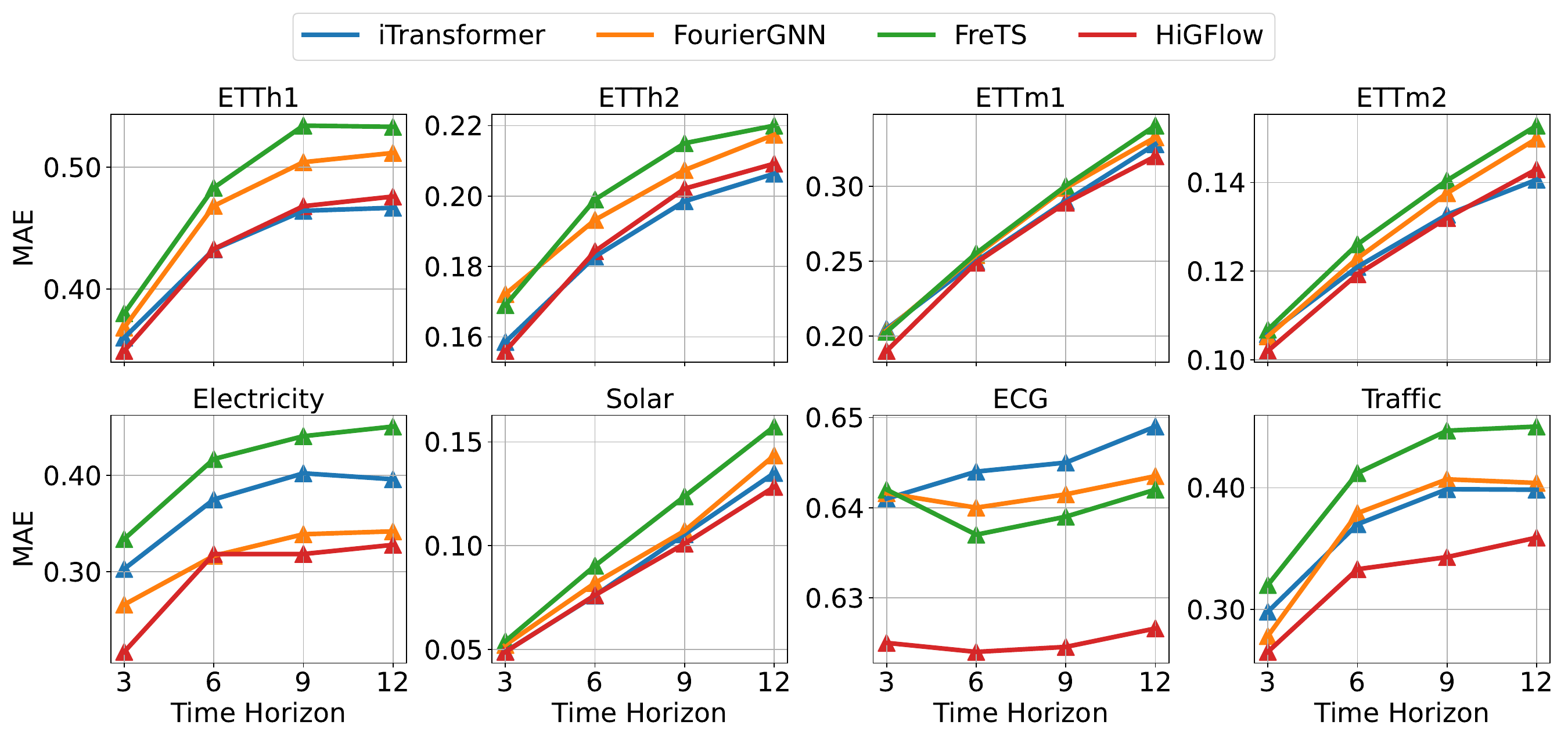}
    \caption{Model performance in terms of MAE while varying prediction buffer size.}
    \label{fig:time_horizon}
\end{figure*}

{\bf Comparison with Baselines.} We evaluate the HiGFlow network against state-of-the-art baselines. Table \ref{tab:results} summarises the experimental results. Compared to non-transformer methods, HiGFlow achieves the lowest MAE and RMSE across all datasets, reducing MAE by 4–18\%, with an average of 7\%, while decreasing RMSE within 2–26\% by a mean of 5\%. Fourier methods, {\it e.g.} FreTS \cite{yi2024frequency}, and Fourier-based GNN methods, such as FourierGNN \cite{yi2024fouriergnn} and StemGNN \cite{cao2020spectral}, excel in capturing spatiotemporal relationships. However, they are restricted to viewing spatial-temporal relationships at a single resolution. Therefore, while the single-depth model HiGFlow(1), which incorporates elements from these methods, often achieves comparable accuracy, the multi-depth HiGFlow(*) model outperforms these models by obtaining lower error metrics.

Table \ref{tab:results} also compares HiGFlow with transformer baselines. While these models do not explicitly capture joint spatial-temporal correlations, they are excellent at extracting relationships along the temporal dimension. HiGFlow outperforms transformers such as Autoformer \cite{wu2021autoformer} and Informer \cite{zhou2021informer} across all datasets, achieving an average of at least 15\% and 9\% improvement in MAE and RMSE. However, compared to state-of-the-art forecasting models like iTransformer and iFlashformer \cite{liu2023itransformer}, HiGFlow shows at worst only marginal gains on the ETT dataset series. In contrast, across the first four datasets, HiGFlow achieves more substantial improvements; on average 12\% and 8\% in MAE and 11\% and 7\% in RMSE for iTransformer and iFlashformer, respectively. This underscores an area of key benefit for HiGFlow, and a relative limitation of transformers. These small dataset have a limited degree of spatial correlation, making dependence on temporal variations more pronounced. Therefore comparative performance between state-of-the-art transformer models and HiGFlow is higher than other other datasets with more variables. 

{\bf Increasing Prediction Buffer Size.} Figure \ref{fig:time_horizon} illustrates the effect of varying the prediction buffer size from 3 to 12 on MAE for HiGFlow, FourierGNN \cite{yi2024fouriergnn}, FreTS \cite{yi2024frequency} and iTransformer \cite{liu2023itransformer}. HiGFlow generally outperforms across all datasets. Exceptions include comparable performance with FourierGNN on Electricity when the buffer size is 6, iTransformer achieving lower error metrics when the time buffer exceeds 6 on ETTh1 and ETTh2, and when the buffer is 12 on ETTm2. Non-transformer baseline models are Fourier-based, and therefore excel in extrapolation over long lead times, however, in some datasets, such as Traffic, Solar, and ETTh1, HiGFlow's relative gain increases with prediction buffer size. The average percentage gain across all datasets from a lead time of 6 onwards is 4.85\%, 5.46\% and 5.56\%. Similar to Table \ref{tab:results}, HiGFlow and the transformer baseline often show comparable performance on the ETT dataset, but HiGFlow exhibits a clear advantage on the remaining datasets.

\begin{figure*}[!htbp]
    \centering    
    \includegraphics[width=1.825\columnwidth]{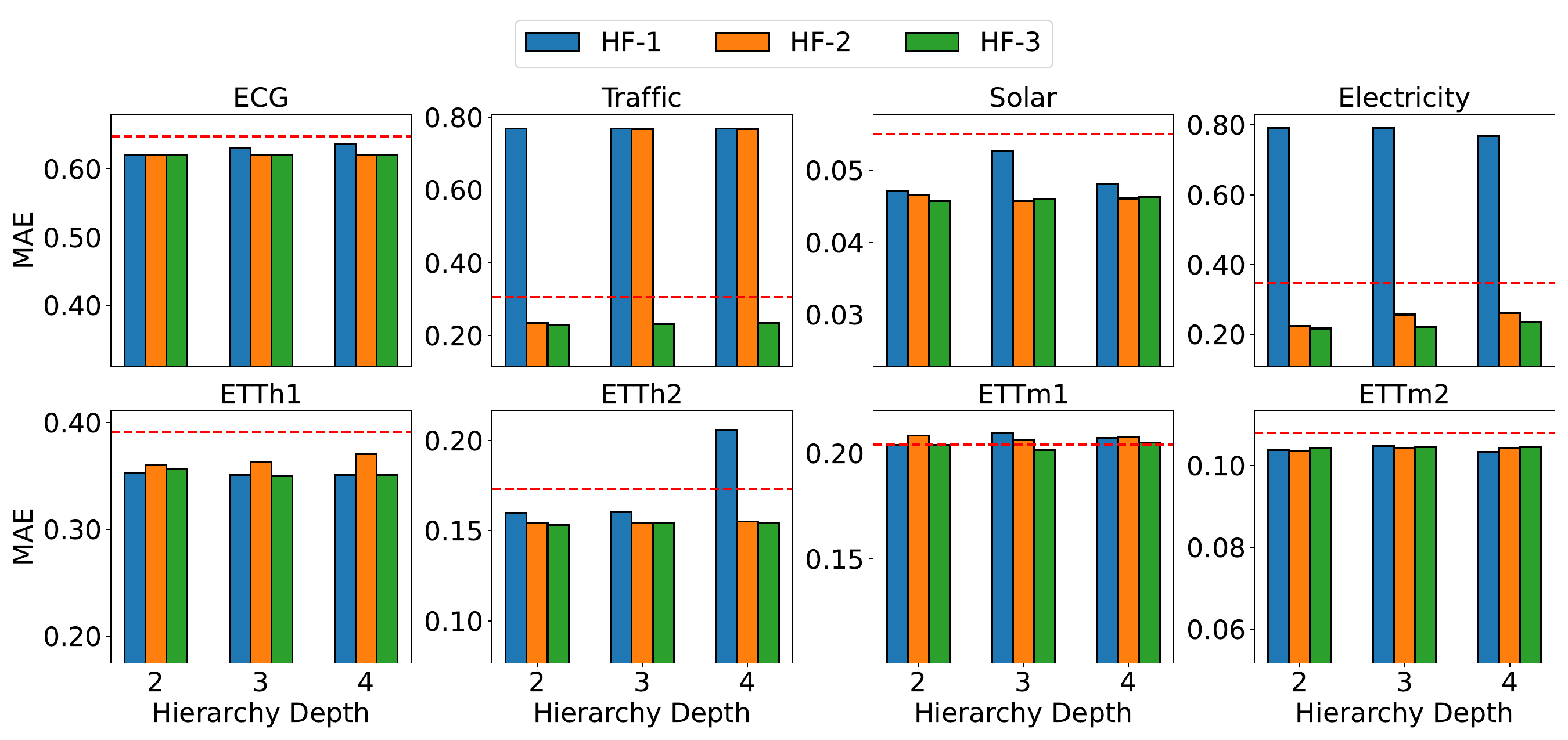}
    \caption{Mean Absolute Error (MAE) as a function of hierarchy depth, showing the impact of varying the non-linearity of the embedding and lifting networks in HiGFlow.}
    \label{fig:depth}
\end{figure*}

{\bf Varying Hierarchical Depth.} We analyse the joint relationship between hierarchical depth, the non-linearity of transition functions in embedding and lifting networks, and the performance of HiGFlow. We report MAE, which also doubles as a proxy for feature space smoothness \cite{rusch2023survey}, and further demonstrates Theorems \ref{th:th1} and \ref{th:th2} in a real-world setting. Specifically, we compare the relative performance of four variations: HiGFlow with strictly linear transition functions, \textbf{HF-1}, which uses matrix multiplication (as in \citet{cini2023graph}) to map clusters to super-nodes; a depth-two network with one hidden layer, \textbf{HF-2}; a depth-three network with two hidden layers, \textbf{HF-3}; and the single-depth HiGFlow model in Table \ref{tab:results}, called {\bf HiGFlow(1)}. Recall that, by the universality theorem \cite{hornik1989multilayer}, both HF-2 and HF-3 are considered universal approximators. Figure~\ref{fig:depth} presents these experiments.

In a setting where hierarchical depth is two, we observe that HF-1 generally achieves comparable performance to HF-2 and HF-3 across most datasets. Exceptions include Traffic and Electricity, where MAE increases substantially beyond HiGFlow(1), demonstrating the impact of information loss from a single round of transition function mappings on large datasets with complex spatiotemporal relationships. However, with greater hierarchical depth, HF-1 shows substantial increases in MAE on several datasets; ECG, ETTh2, ETTm1, and Solar. For ETTh2 and ETTm1, this greater than HiGFlow(1). An analogous statement can be made for HF-2 on Traffic and ETTh1, though performance on all other datasets tends to be more stable than HF-1. 

Our results show that non-linearity is essential for achieving hierarchical depth, which aligns with our theoretical analysis. However, a linear transition function proves sufficient when the feature space distribution has limited support, as observed in the ETTh1 and ETTm2 datasets. In contrast, datasets like Traffic and Electricity demonstrate the necessity of non-linearity, where substantial increases in error metrics emerge even at a depth of two.

{\bf Parameters and Runtime Analysis.} We evaluate the computational overhead of HiGFlow by analysing its training runtime and memory cost relative to the number of parameters. Table \ref{tab:training_time_parameters} summarises the results, showing the impact of varying HiGFlow's depth on memory usage and runtime across the ETTh1 and ECG datasets. Notably, HiGFlow exhibits a lower parameter count relative to all other baselines. Compared to FourierGNN \cite{yi2024fouriergnn}, which leverages temporal embeddings, HiGFlow achieves a reduction of 7.2\% and 6.5\% on ETTh1 and ECG, respectively. However, this efficiency diminishes as dataset size increases, with reductions of 7.2\%, 7.1\%, 6.9\%, and 6.5\% on ETTh1 compared to 6.5\%, 6.3\%, 5.9\%, and 5.7\% on ECG.

During the forward pass, certain operations in HiGFlow, such as finding the strongly connected components of a graph, are not easily vectorisable. This limitation, combined with existing overheads, results in longer training time, at least 1.5 times higher than other baselines. Additionally, this computational cost grows significantly with increased depth, scaling by factors of 2.3, 3.4, and 4.1, respectively.

\begin{table*}[ht]
    \centering
    \fontsize{8pt}{8pt}\selectfont
    \caption{Training Time and Parameters for Various Models}
    \begin{tabular}{lcccc}
        \toprule
        & \multicolumn{2}{c}{ETTh1} & \multicolumn{2}{c}{ECG} \\
        \cmidrule(lr){2-3} \cmidrule(lr){4-5}
         & Parameters & Training (s/epoch) & Parameters & Training (s/epoch) \\
        \midrule
        StemGNN & $194,294$ & $4.10 \pm 0.25$ & $258,799$ & $1.75\pm0.19$ \\
        FreTS  & $460,419$ & $9.33 \pm 0.67$ & $460,419$ & $0.66\pm0.16$ \\
        FourierGNN & $182,307$ & $2.52 \pm 0.36$ & $182,307$  & $0.84\pm0.01$ \\
        iTransformer & $6,313,987$ & $3.50\pm0.63$ & $6,313,987$ & $2.56\pm0.47$ \\
        \midrule
        HiGFlow(1) & $30,291$ & $3.25 \pm 0.29$ & $30,291$ & $2.74\pm0.29$ \\
        HiGFlow(2) & $85,531$ & $7.44 \pm 0.38$ & $160,142$ & $8.81\pm0.32$ \\
        HiGFlow(3) & $139,483$ & $11.69 \pm 0.25$ & $265,178$ & $13.04\pm0.32$ \\
        HiGFlow(4) & $192,596$ & $16.21 \pm 0.52$ & $355,821$ & $16.03\pm0.31$ \\
        \bottomrule
    \end{tabular}
    \label{tab:training_time_parameters}
\end{table*}
\begin{table*}[!htbp]
    \fontsize{8pt}{10pt}\selectfont
        \centering
        \caption{Ablation study on ETTh1, ECG, and Traffic datasets. Results are reported for MAE and RMSE.}
        \begin{tabular}{lcccccccc}
        \toprule
        Dataset & Metric & No/Embed-Domain & No/Lift-Domain & No/Lift-Embed & No/Domain-Embed & No/Embed & HiGFlow \\ 
        \midrule
        \multirow{2}{*}{ETTh1} 
        & MAE  & 0.353 & 0.352 & {\bf 0.345} & 0.353 & 0.351 & 0.349 \\
        & RMSE & 0.539 & 0.538 & 0.535 & 0.543 & 0.538 & {\bf 0.532} \\
        \midrule
        \multirow{2}{*}{ECG} 
        & MAE  & 0.623 & 0.623 & 0.624 & 0.624 & 0.624 & {\bf 0.621} \\
        & RMSE & 0.892 & 0.892 & 0.892 & 0.892 & 0.892 & {\bf 0.889} \\
        \midrule
        \multirow{2}{*}{Traffic} 
        & MAE  & 0.264    & 0.268    & {\bf 0.253}    & 0.264    & 0.273    & 0.265    \\ 
        & RMSE & 0.537    & 0.538   & {\bf 0.520}    & 0.536    & 0.545    & 0.534    \\
        \bottomrule
    \end{tabular}
    \label{table:ablation}
\end{table*}

{\bf Ablation Test.} We conduct an ablation study to evaluate the empirical utility of embeddings in the Lifter and Embedding networks (Sections \ref{sec:lifting} and \ref{sec:embedding}) on the ETTh1, ECG and Traffic datasets. Table \ref{table:ablation} presents results comparing the HiGFlow network with configurations excluding node embeddings in the embedding network ({\bf No/Embed-Domain}) or lifting network ({\bf No/Lift-Domain}). Additional baselines include the absence of both lifting network domain and image shifts ({\bf No/Lift-Embed}), domain embeddings in either network ({\bf No/Domain-Embed}), or embeddings altogether ({\bf No/Embed}). The results indicate that incorporating domain shifts in the embedding network consistently improves performance, as seen in the performance of No/Lift-Embed relative to No/Embed. However, to benefit from node embeddings in the domain of the lifting network, the na\"ive-encoding (image-shift) must also included, as demonstrated by comparing HiGFlow and No/Lift-Embed with No/Lift-Domain. Even then, performance relative to only embedding network domain shifts may degrade in some instances, such as with the Traffic dataset.

\section{Related Work}

We review recent advancements in time series forecasting and hierarchical modelling.

\subsection{Time Series Forecasting}

{\bf Time-series to Graph Embeddings.} \citet{gao2022equivalence} prove theoretically that message-passing neural networks trained on GRU-embedded feature spaces are equivalent in expressivity to the Weisfeiler-Lehman (WL) test. \citet{cao2020spectral} propose a parameter-efficient method for embedding time-series data into graphs by mapping variables onto nodes, effectively capturing implicit temporal structures. In contrast, \citet{yi2024fouriergnn} consider both spatial and temporal interdependencies by constructing a hypervariate graph, where each variable at any time step is represented as a unique node and subsequently applying graph convolutions in Fourier space.

{\bf Fourier Methods for Forecasting.} Fourier methods effectively transform temporal dimensions into scalar frequency values, serving as a form of attention mechanism \cite{yi2024frequency}. Traditionally, these methods truncate high-frequency components beyond a specific threshold \cite{li2020fourier}. Recent studies, however, demonstrate that retaining high-frequency components improves the forecast's high-correlation lead time, resulting in more accurate long-term predictions \cite{lippe2023pde, zhou2022fedformer}.

\subsection{Hierarchical Graph Methods}

{\bf Pooling.} Graph pooling methods aim to coarsen a graph, abstracting cliques of nodes onto super node representations \cite{lee2019self, zhang2019hierarchical}. Hierarchical pooling \cite{ying2018hierarchical} offers a straightforward method to reduce granularity, allowing for tailored preservation of high-frequency components \cite{bianchi2023expressive}. 

{\bf  Hierarchical Forecasting.} Early methods \cite{wang2018multilevel} exploit relational structures by progressively building predictions from the lowest resolution level. 
Recent works focus on global or regional forecasting, such as \cite{oskarsson2024probabilistic}, employ non-homogeneous MPNNs to map between predefined mesh graphs. However, these techniques rely on apriori knowledge of graph structure.  

\section{Conclusion} 

We investigated the problem of information retention in hierarchical forecasting models, and proposed the HiGFlow network, which stores mixed-resolution information in a memory buffer persistent across all hierarchical levels. HiGFlow thereby enables consideration of spatiotemporal relationships at multiple resolutions. We prove three key theorems, demonstrating that HiGFlow reduces the smoothness of embedded and lifted spaces, and furthermore non-strictly increases WL expressivity as hierarchical layers are stacked. Our empirical evaluation corroborates this in practice and further highlights HiGFlow's ability to achieve lower MAE and RMSE relative to existing state-of-the-art graphical forecasting and transformer baselines.


\bibliography{paper}

\begin{thebibliography}{40}
\providecommand{\natexlab}[1]{#1}
\providecommand{\url}[1]{\texttt{#1}}
\expandafter\ifx\csname urlstyle\endcsname\relax
  \providecommand{\doi}[1]{doi: #1}\else
  \providecommand{\doi}{doi: \begingroup \urlstyle{rm}\Url}\fi

\bibitem[Alon \& Yahav(2021)Alon and Yahav]{alon2020bottleneck}
Alon, U. and Yahav, E.
\newblock On the bottleneck of graph neural networks and its practical implications.
\newblock \emph{International Conference on Learning Representations}, 2021.

\bibitem[Bauer et~al.(2015)Bauer, Thorpe, and Brunet]{bauer2015quiet}
Bauer, P., Thorpe, A., and Brunet, G.
\newblock The quiet revolution of numerical weather prediction.
\newblock \emph{Nature}, 525\penalty0 (7567):\penalty0 47--55, 2015.

\bibitem[Bianchi \& Lachi(2023)Bianchi and Lachi]{bianchi2023expressive}
Bianchi, F.~M. and Lachi, V.
\newblock The expressive power of pooling in graph neural networks.
\newblock \emph{Advances in neural information processing systems}, 36:\penalty0 71603--71618, 2023.

\bibitem[Cao et~al.(2020)Cao, Wang, Duan, Zhang, Zhu, Huang, Tong, Xu, Bai, Tong, et~al.]{cao2020spectral}
Cao, D., Wang, Y., Duan, J., Zhang, C., Zhu, X., Huang, C., Tong, Y., Xu, B., Bai, J., Tong, J., et~al.
\newblock Spectral temporal graph neural network for multivariate time-series forecasting.
\newblock \emph{Advances in Neural Information Processing Systems}, 33:\penalty0 17766--17778, 2020.

\bibitem[Chung et~al.(2014)Chung, Gulcehre, Cho, and Bengio]{chung2014empirical}
Chung, J., Gulcehre, C., Cho, K., and Bengio, Y.
\newblock Empirical evaluation of gated recurrent neural networks on sequence modeling.
\newblock \emph{Advances in Neural Information Processing Systems}, 2014.

\bibitem[Cini et~al.(2023)Cini, Mandic, and Alippi]{cini2023graph}
Cini, A., Mandic, D., and Alippi, C.
\newblock Graph-based time series clustering for end-to-end hierarchical forecasting.
\newblock \emph{International Conference on Learning Representations}, 2023.

\bibitem[Cuomo et~al.(2022)Cuomo, Di~Cola, Giampaolo, Rozza, Raissi, and Piccialli]{cuomo2022scientific}
Cuomo, S., Di~Cola, V.~S., Giampaolo, F., Rozza, G., Raissi, M., and Piccialli, F.
\newblock Scientific machine learning through physics--informed neural networks: Where we are and what’s next.
\newblock \emph{Journal of Scientific Computing}, 92\penalty0 (3):\penalty0 88, 2022.

\bibitem[Dhillon et~al.(2007)Dhillon, Guan, and Kulis]{dhillon2007weighted}
Dhillon, I.~S., Guan, Y., and Kulis, B.
\newblock Weighted graph cuts without eigenvectors a multilevel approach.
\newblock \emph{IEEE transactions on pattern analysis and machine intelligence}, 29\penalty0 (11):\penalty0 1944--1957, 2007.

\bibitem[Di~Giovanni et~al.(2023)Di~Giovanni, Giusti, Barbero, Luise, Lio, and Bronstein]{di2023over}
Di~Giovanni, F., Giusti, L., Barbero, F., Luise, G., Lio, P., and Bronstein, M.~M.
\newblock On over-squashing in message passing neural networks: The impact of width, depth, and topology.
\newblock In \emph{International Conference on Machine Learning}, pp.\  7865--7885. PMLR, 2023.

\bibitem[Gao \& Ribeiro(2022)Gao and Ribeiro]{gao2022equivalence}
Gao, J. and Ribeiro, B.
\newblock On the equivalence between temporal and static equivariant graph representations.
\newblock In \emph{International Conference on Machine Learning}, pp.\  7052--7076. PMLR, 2022.

\bibitem[Guo et~al.(2021)Guo, Hu, Sun, Qian, Gao, and Yin]{guo2021hierarchical}
Guo, K., Hu, Y., Sun, Y., Qian, S., Gao, J., and Yin, B.
\newblock Hierarchical graph convolution network for traffic forecasting.
\newblock In \emph{Proceedings of the AAAI conference on artificial intelligence}, volume~35, pp.\  151--159, 2021.

\bibitem[Hamilton et~al.(2017)Hamilton, Ying, and Leskovec]{hamilton2017inductive}
Hamilton, W., Ying, Z., and Leskovec, J.
\newblock Inductive representation learning on large graphs.
\newblock In \emph{Advances in Neural Information Processing Systems}, 2017.

\bibitem[Hochreiter(1997)]{hochreiter1997long}
Hochreiter, S.
\newblock Long short-term memory.
\newblock \emph{Neural Computation MIT-Press}, 1997.

\bibitem[Hornik et~al.(1989)Hornik, Stinchcombe, and White]{hornik1989multilayer}
Hornik, K., Stinchcombe, M., and White, H.
\newblock Multilayer feedforward networks are universal approximators.
\newblock \emph{Neural networks}, 2\penalty0 (5):\penalty0 359--366, 1989.

\bibitem[Hua et~al.(2022)Hua, Dai, Liu, and Le]{hua2022transformer}
Hua, W., Dai, Z., Liu, H., and Le, Q.
\newblock Transformer quality in linear time.
\newblock In \emph{International conference on machine learning}, pp.\  9099--9117. PMLR, 2022.

\bibitem[Karniadakis et~al.(2021)Karniadakis, Kevrekidis, Lu, Perdikaris, Wang, and Yang]{karniadakis2021physics}
Karniadakis, G.~E., Kevrekidis, I.~G., Lu, L., Perdikaris, P., Wang, S., and Yang, L.
\newblock Physics-informed machine learning.
\newblock \emph{Nature Reviews Physics}, 3\penalty0 (6):\penalty0 422--440, 2021.

\bibitem[Kipf \& Welling(2016)Kipf and Welling]{kipf2016semi}
Kipf, T.~N. and Welling, M.
\newblock Semi-supervised classification with graph convolutional networks.
\newblock \emph{International Conference on Learning Representations}, 2016.

\bibitem[Lam et~al.(2023)Lam, Sanchez-Gonzalez, Willson, Wirnsberger, Fortunato, Alet, Ravuri, Ewalds, Eaton-Rosen, Hu, Merose, Hoyer, Holland, Vinyals, Stott, Pritzel, Mohamed, and Battaglia]{lam2023graphcast}
Lam, R., Sanchez-Gonzalez, A., Willson, M., Wirnsberger, P., Fortunato, M., Alet, F., Ravuri, S., Ewalds, T., Eaton-Rosen, Z., Hu, W., Merose, A., Hoyer, S., Holland, G., Vinyals, O., Stott, J., Pritzel, A., Mohamed, S., and Battaglia, P.
\newblock Learning skillful medium-range global weather forecasting.
\newblock \emph{Science}, 382\penalty0 (6677):\penalty0 1416--1421, 2023.
\newblock \doi{10.1126/science.adi2336}.

\bibitem[Lee et~al.(2019)Lee, Lee, and Kang]{lee2019self}
Lee, J., Lee, I., and Kang, J.
\newblock Self-attention graph pooling.
\newblock In \emph{International conference on Machine Learning}, pp.\  3734--3743. pmlr, 2019.

\bibitem[Li et~al.(2020)Li, Kovachki, Azizzadenesheli, Liu, Bhattacharya, Stuart, and Anandkumar]{li2020fourier}
Li, Z., Kovachki, N., Azizzadenesheli, K., Liu, B., Bhattacharya, K., Stuart, A., and Anandkumar, A.
\newblock Fourier neural operator for parametric partial differential equations.
\newblock \emph{arXiv preprint arXiv:2010.08895}, 2020.

\bibitem[Lippe et~al.(2023)Lippe, Veeling, Perdikaris, Turner, and Brandstetter]{lippe2023pde}
Lippe, P., Veeling, B., Perdikaris, P., Turner, R., and Brandstetter, J.
\newblock Pde-refiner: Achieving accurate long rollouts with neural pde solvers.
\newblock \emph{Advances in Neural Information Processing Systems}, 36:\penalty0 67398--67433, 2023.

\bibitem[Liu et~al.(2024)Liu, Hu, Zhang, Wu, Wang, Ma, and Long]{liu2023itransformer}
Liu, Y., Hu, T., Zhang, H., Wu, H., Wang, S., Ma, L., and Long, M.
\newblock itransformer: Inverted transformers are effective for time series forecasting.
\newblock \emph{International Conference on Learning Representations}, 2024.

\bibitem[Nie et~al.(2023)Nie, Nguyen, Sinthong, and Kalagnanam]{nie2022time}
Nie, Y., Nguyen, N.~H., Sinthong, P., and Kalagnanam, J.
\newblock A time series is worth 64 words: Long-term forecasting with transformers.
\newblock \emph{International Conference on Learning Representations}, 2023.

\bibitem[Oskarsson et~al.(2024)Oskarsson, Landelius, Deisenroth, and Lindsten]{oskarsson2024probabilistic}
Oskarsson, J., Landelius, T., Deisenroth, M.~P., and Lindsten, F.
\newblock Probabilistic weather forecasting with hierarchical graph neural networks.
\newblock In \emph{The Thirty-eighth Annual Conference on Neural Information Processing Systems}, 2024.
\newblock URL \url{https://openreview.net/forum?id=wTIzpqX121}.

\bibitem[Rusch et~al.(2023)Rusch, Bronstein, and Mishra]{rusch2023survey}
Rusch, T.~K., Bronstein, M.~M., and Mishra, S.
\newblock A survey on oversmoothing in graph neural networks.
\newblock \emph{arXiv preprint arXiv:2303.10993}, 2023.

\bibitem[Vaswani(2017)]{vaswani2017attention}
Vaswani, A.
\newblock Attention is all you need.
\newblock \emph{Advances in Neural Information Processing Systems}, 2017.

\bibitem[Veli{\v{c}}kovi{\'c} et~al.(2018)Veli{\v{c}}kovi{\'c}, Cucurull, Casanova, Romero, Li{\`o}, and Bengio]{velickovic2018graph}
Veli{\v{c}}kovi{\'c}, P., Cucurull, G., Casanova, A., Romero, A., Li{\`o}, P., and Bengio, Y.
\newblock Graph attention networks.
\newblock In \emph{International Conference on Learning Representations}, 2018.

\bibitem[Wang et~al.(2018)Wang, Wang, Li, and Wu]{wang2018multilevel}
Wang, J., Wang, Z., Li, J., and Wu, J.
\newblock Multilevel wavelet decomposition network for interpretable time series analysis.
\newblock In \emph{Proceedings of the 24th ACM SIGKDD International Conference on Knowledge Discovery and Data Mining}, pp.\  2437--2446, 2018.

\bibitem[Wu et~al.(2021)Wu, Xu, Wang, and Long]{wu2021autoformer}
Wu, H., Xu, J., Wang, J., and Long, M.
\newblock Autoformer: Decomposition transformers with auto-correlation for long-term series forecasting.
\newblock \emph{Advances in neural information processing systems}, 34:\penalty0 22419--22430, 2021.

\bibitem[Wu et~al.(2022)Wu, Cui, Pei, Zhao, and Guo]{wu2022graph}
Wu, L., Cui, P., Pei, J., Zhao, L., and Guo, X.
\newblock Graph neural networks: foundation, frontiers and applications.
\newblock In \emph{Proceedings of the 28th ACM SIGKDD Conference on Knowledge Discovery and Data Mining}, pp.\  4840--4841, 2022.

\bibitem[Xiong et~al.(2021)Xiong, Xiong, Chen, Jiang, and Zheng]{xiong2021graph}
Xiong, J., Xiong, Z., Chen, K., Jiang, H., and Zheng, M.
\newblock Graph neural networks for automated de novo drug design.
\newblock \emph{Drug discovery today}, 26\penalty0 (6):\penalty0 1382--1393, 2021.

\bibitem[Xu et~al.(2018)Xu, Hu, Leskovec, and Jegelka]{xu2018powerful}
Xu, K., Hu, W., Leskovec, J., and Jegelka, S.
\newblock How powerful are graph neural networks?
\newblock \emph{International Conference on Learning Representations}, 2018.

\bibitem[Yang et~al.(2021)Yang, Zhao, Rong, Yan, Li, Ma, and Huang]{yang2021hierarchical}
Yang, J., Zhao, P., Rong, Y., Yan, C., Li, C., Ma, H., and Huang, J.
\newblock Hierarchical graph capsule network.
\newblock In \emph{Proceedings of the AAAI Conference on Artificial Intelligence}, volume~35, pp.\  10603--10611, 2021.

\bibitem[Yi et~al.(2024{\natexlab{a}})Yi, Zhang, Fan, He, Hu, Wang, An, Cao, and Niu]{yi2024fouriergnn}
Yi, K., Zhang, Q., Fan, W., He, H., Hu, L., Wang, P., An, N., Cao, L., and Niu, Z.
\newblock Fouriergnn: Rethinking multivariate time series forecasting from a pure graph perspective.
\newblock \emph{Advances in Neural Information Processing Systems}, 36, 2024{\natexlab{a}}.

\bibitem[Yi et~al.(2024{\natexlab{b}})Yi, Zhang, Fan, Wang, Wang, He, An, Lian, Cao, and Niu]{yi2024frequency}
Yi, K., Zhang, Q., Fan, W., Wang, S., Wang, P., He, H., An, N., Lian, D., Cao, L., and Niu, Z.
\newblock Frequency-domain mlps are more effective learners in time series forecasting.
\newblock \emph{Advances in Neural Information Processing Systems}, 36, 2024{\natexlab{b}}.

\bibitem[Ying et~al.(2018)Ying, You, Morris, Ren, Hamilton, and Leskovec]{ying2018hierarchical}
Ying, Z., You, J., Morris, C., Ren, X., Hamilton, W., and Leskovec, J.
\newblock Hierarchical graph representation learning with differentiable pooling.
\newblock \emph{Advances in neural information processing systems}, 31, 2018.

\bibitem[Zhang et~al.(2017)Zhang, Aggarwal, and Qi]{zhang2017stock}
Zhang, L., Aggarwal, C., and Qi, G.-J.
\newblock Stock price prediction via discovering multi-frequency trading patterns.
\newblock In \emph{Proceedings of the 23rd ACM SIGKDD International Conference on Knowledge Discovery and Data Mining}, pp.\  2141--2149, 2017.

\bibitem[Zhang et~al.(2019)Zhang, Bu, Ester, Zhang, Yao, Yu, and Wang]{zhang2019hierarchical}
Zhang, Z., Bu, J., Ester, M., Zhang, J., Yao, C., Yu, Z., and Wang, C.
\newblock Hierarchical graph pooling with structure learning.
\newblock \emph{Association for the Advancement of Artificial Intelligence}, 2019.

\bibitem[Zhou et~al.(2021)Zhou, Zhang, Peng, Zhang, Li, Xiong, and Zhang]{zhou2021informer}
Zhou, H., Zhang, S., Peng, J., Zhang, S., Li, J., Xiong, H., and Zhang, W.
\newblock Informer: Beyond efficient transformer for long sequence time-series forecasting.
\newblock In \emph{Proceedings of the AAAI conference on artificial intelligence}, volume~35, pp.\  11106--11115, 2021.

\bibitem[Zhou et~al.(2022)Zhou, Ma, Wen, Wang, Sun, and Jin]{zhou2022fedformer}
Zhou, T., Ma, Z., Wen, Q., Wang, X., Sun, L., and Jin, R.
\newblock Fedformer: Frequency enhanced decomposed transformer for long-term series forecasting.
\newblock In \emph{International conference on machine learning}, pp.\  27268--27286. PMLR, 2022.

\end{thebibliography}
\bibliographystyle{icml2024}

\end{document}